%% file: acl2019.tex
\documentclass[11pt,a4paper]{article}
\usepackage[hyperref]{acl2019}
\usepackage{times}
\usepackage{latexsym}
\usepackage{amssymb}
\usepackage{amsmath} 
\usepackage{booktabs}
\usepackage{enumerate}
\usepackage{graphicx}
\usepackage{xspace}
\usepackage{float}
\usepackage{bbm}
\usepackage{bm}
\usepackage{multirow}
\usepackage{booktabs}
\usepackage{color}
\usepackage{framed}
\usepackage{stfloats}
\usepackage[T1]{fontenc}
\definecolor{shadecolor}{RGB}{180,180,180}
\usepackage{url}
\newcommand{\paratitle}[1]{\vspace{1.5ex}\noindent\textbf{#1}}
\newcommand{\ie}{\emph{i.e.,}\xspace}

\newcommand{\eg}{\emph{e.g.,}\xspace}

\newcommand{\ignore}[1]{}
\newcommand{\tabincell}[2]{\begin{tabular}{@{}#1@{}}#2\end{tabular}}

\aclfinalcopy 
\def\aclpaperid{151} 

\title{Generating Long and Informative Reviews with Aspect-Aware Coarse-to-Fine Decoding}

\author{
	Junyi Li\textsuperscript{\rm{1}}, 
	Wayne Xin Zhao\textsuperscript{\rm{1,2\thanks{\ \ Corresponding author}}}, 
	Ji-Rong Wen\textsuperscript{\rm{1,2}},
	{\rm and} Yang Song\textsuperscript{\rm{3}} \\
	\textsuperscript{1}School of Information, Renmin University of China \\
	\textsuperscript{2}Beijing Key Laboratory of Big Data Management and Analysis Methods \\
	\textsuperscript{3}Boss Zhipin \\
	\texttt{\{lijunyi,jrwen\}@ruc.edu.cn} \\ \texttt{batmanfly@gmail.com} \quad \texttt{songyang@kanzhun.com}\\
}

\date{}

\begin{document}
\captionsetup[figure]{labelsep=period}
\captionsetup[table]{labelsep=period}
\maketitle
\begin{abstract}
Generating long and informative review text is a challenging natural language generation task. 
Previous work focuses on word-level generation, neglecting the importance of topical and syntactic characteristics from natural languages.
In this paper, we propose a novel review generation model by characterizing an elaborately designed aspect-aware coarse-to-fine generation process. First, we model the aspect transitions to capture the overall content flow. Then, to generate a sentence, an aspect-aware sketch will be predicted using an aspect-aware decoder. Finally, another decoder fills in the semantic slots by generating corresponding words.
Our approach is able to jointly utilize aspect semantics, syntactic sketch, and context information.
Extensive experiments results have demonstrated the effectiveness of the proposed model.
\end{abstract}

\input{sec-intro}

\input{sec-rel}

\input{sec-model}

\input{sec-exp}

\input{sec-con}

\input{sec-ack}

\bibliography{acl2019}
\bibliographystyle{acl_natbib}

\end{document}

%% file: sec-intro.tex
\section{Introduction}
In the past decades, online review services (\eg \textsc{Amazon} and \textsc{Yelp}) have been an important kind of information platforms where users post their feedbacks or comments about products~\cite{kim2016analysis}.
Usually, writing an informative and well-structured review will require considerable efforts by users. To assist the writing process, the task of review generation has been proposed to automatically generate review text for a user given a product and her/his rating on it~\cite{TangYCZM16,ZhouLWDHX17}.

In the literature, various methods have been developed for review generation~\cite{TangYCZM16,ZhouLWDHX17,NiLVM17,WangZ17,Catherine2018}.
Most of these methods adopt Recurrent Neural Networks (RNN) based methods, especially the improved variants of Long-Short Term Memory~(LSTM)~\cite{HochreiterS97} and Gated Recurrent Unit~(GRU)~\cite{ChoMGBBSB14}. They fulfill the review generation task by performing the decoding conditioned on useful context information. 
Usually, an informative review is likely to consist of multiple sentences, containing substantive comments from users.
Hence, a major problem of existing RNN-based methods is that they have limited capacities in producing long and informative text. 
More recently, Generative Adversarial Net (GAN) based methods~\cite{ZangW17,YuZWY17,GuoLCZYW18,XuRL018} have been proposed 
to enhance the generation of long, diverse and novel text. However, they still 
focus on word-level generation, and neglect the importance of topical and syntactic characteristics from natural languages. 

\begin{figure}[tb]
\centering 
\includegraphics[width=0.4\textwidth]{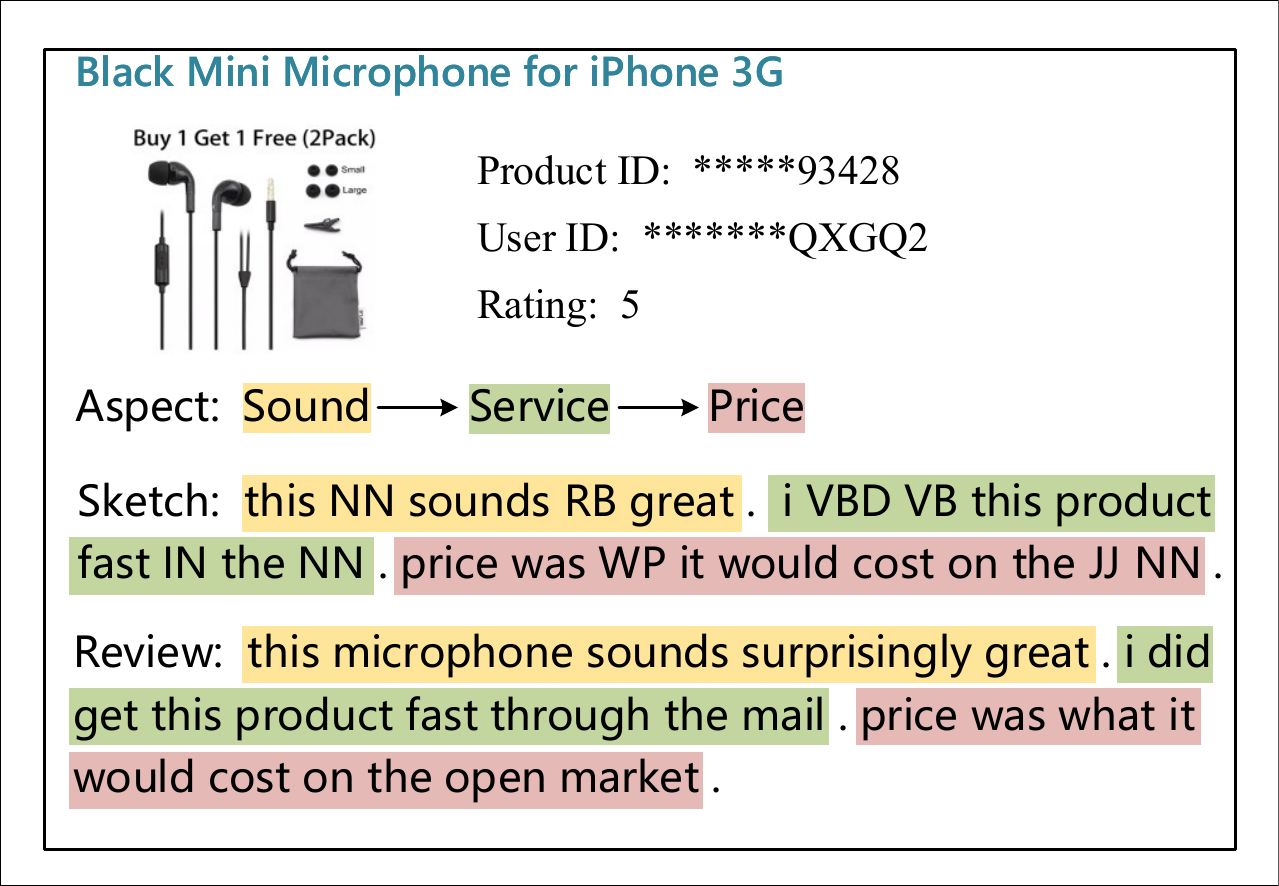} 
\caption{An illustrative example for our generation process. We select a sample review on \textsc{Amazon}. The aspect labels and sketches are manually created for explaining our idea, which will be learned by our model. } 
\label{fig-example} 
\vspace{-0.3cm}
\end{figure}

As found in the literature of linguistics~\cite{pullum2010} and writing~\cite{bateman2003natural},  the writing process itself has involved multiple stages focusing on different levels of goals. We argue that an ideal review generation approach should follow the writing procedure of a real user and capture rich characteristics from natural language. 
With this motivation, we design an elaborative coarse-to-fine generation process by considering the aspect semantics and syntactic characteristics.
 Figure~\ref{fig-example} presents an illustrative example for our review generation process.  First, we conceive the content flow that is characterized as an aspect sequence. An aspect describes some property or attribute about a product~\cite{Zhao-emnlp-2010}, such as \emph{sound} and \emph{service} in this example.
To generate a sentence, we further create a sentence skeleton containing semantic slots given the aspect semantics. The semantic slots denote the placeholders for useful syntactic information~(\eg Part-of-speech tags). Finally, the semantic slots are filled with the generated words. 
The process is repeated until all sentences are generated. 


Based on such a generation process, in this paper, we propose a novel aspect-aware coarse-to-fine decoder for generating product reviews.  We first utilize unsupervised topic models to extract aspects and tag review sentences with aspect labels. 
We develop an attention-based RNN decoder to generate the aspect sequence conditioned on the context including users, items and ratings. 
By modeling  the transitions of aspect semantics among sentences, we are able to capture the content flow of the whole review. 
Then, we generate a  semantic template called \emph{sketch} using an aspect-aware decoder, which represents the sentence skeleton. 
Finally, we generate the word content according to an informed decoder that considers aspect labels, sketch symbols and previously decoded words. 
Extensive experiments  on three real-world review datasets have demonstrated the effectiveness of the proposed model.

To our knowledge, it is the first review generation model that is able to  jointly utilize aspect semantics, syntactic sketch, and context information.
We decompose the entire generation process into three stages. In this way, the generation of long review text becomes more controllable, since we consider a simpler sequence generation task at each stage.
Furthermore, we incorporate language characteristics (\eg Part-of-Speech tags and $n$-grams) into the aspect-aware  decoder to instruct the generation of well-structured text.

%% file: sec-rel.tex
\section{Related Work}

In recent years, researchers have made great progress in natural language generation (NLG) \cite{ChengXGLZF18, ZhaoZWYHZ18, LewisDF18}.  
As a special NLG task, automatic review generation has been proposed to assist the writing of online reviews for users. RNN-based methods have been proposed to generate the review content conditioned on useful context information~\cite{TangYCZM16,ZhouLWDHX17}.
Especially, the task of review generation is closely related to the studies in recommender systems that aim to predict the preference of a user over products.
Hence, several studies propose to couple the solutions of the two lines of research work, and utilize the user-product interactions for improving the review generation~\cite{NiLVM17,WangZ17,Catherine2018,NiM18}. 
Although \citet{NiM18} have explored aspect information to some extent, they characterize the generation process in a single stage and do not perform the coarse-to-fine decoding. 
 Besides, the aspect transition patterns have been not modeled. 

It has been found that RNN models tend to generate short, repetitive, and dull texts~\cite{LinSMS18, LuoXLZ018}. For addressing this issue, Generative Adversarial Nets (GAN) based  approaches have been recently proposed to generate long, diverse and novel text~\cite{ZangW17,YuZWY17,GuoLCZYW18,XuRL018}.
These methods usually utilize reinforcement learning techniques to deal with the generation of  discrete symbols.  However, they seldom consider the linguistic information from natural languages, which cannot fully address the difficulties of our task.

Our work is inspired by the work of using sketches as intermediate representations~\cite{LapataD18, WisemanSR18, XuRZZC018, SuLYC18}.
These works usually focus on sentence- or utterance-level generation tasks, in which global aspect semantics and transitions have not been considered. 
Our work is also related to review data mining, especially the studies on topic or aspect extraction from review data~\cite{QiuYCB17,Zhao-emnlp-2010}.

%% file: sec-model.tex
\section{Problem Formulation}
A review is a natural language text written by a user $u$  on a product (or item) $i$ with a rating score of $r$.
Let  $\mathcal{V}$ denote the vocabulary and $y^{1:m}=\{\langle y_{j,1},\cdots,y_{j,t},\cdots,y_{j,n_j} \rangle \}_{j=1}^m$ denote a review text consisting of $m$ sentences, where $y_{j,t} \in \mathcal{V}$ denotes the $t$-th word  of the $j$-th review sentence and $n_j$ is the length of the $j$-th sentence.  

We assume that the review generation process is decomposed into three different stages.
First, a user generates an aspect sequence representing the major content flow for a review.
To generate a sentence, we predict an aspect-aware sketch conditioned on an aspect label.
Finally, based on the aspect label and the sketch, we generate the word content for a sentence. 
The process is repeated until all the sentences are generated. 

Let $\mathcal{A}$ denote a set of $A$ aspects in our collection. 
Following~\cite{Zhao-emnlp-2010}, we assume each review sentence is associated with an aspect label, describing some property or attribute about a product or an item.
We derive an aspect sequence for a review text, denoted by $a^{1:m}={\langle a_1,\cdots,a_j,\cdots,a_m \rangle}$, where $a_j \in \mathcal{A}$ is the aspect label (or ID) of the $j$-th sentence. 
For each sentence, we assume that it is written according to some semantic sketch, which is also denoted by a symbol sequence. Let $s^{1:m}=\{\langle s_{j,1},\cdots,s_{j,t},\cdots,s_{j,n'_j} \rangle\}_{j=1}^m$, where $n'_j$ is the length of the $j$-th sketch, and $s_{j,t}$ is the $t$-th token of the $j$-th sketch denoting a word, a Part-of-Speech tag, a bi-gram, etc.

Based on the above notations, we are ready to define our task.  
Given user $u$,  item $i$ and the rating score $r$, we aim to automatically generate a review that is able to maximize the joint probability of the aspects, sketches and words

\begin{small}
\begin{eqnarray}\label{eq-joint}
&&\text{Pr}(y^{1:m},  s^{1:m}, a^{1:m} |c)\\\nonumber
&=& \text{Pr}(a^{1:m}|c) \text{Pr}(s^{1:m}|a^{1:m},c) \text{Pr}(y^{1:m}|a^{1:m},s^{1:m},c),\\\nonumber
&=& \prod_{j=1}^m \text{Pr}(a_j|a_{<j},c) \prod_{j,t} \text{Pr}(s_{j,t} | s_{j,<t}, a_j,c)\\\nonumber
&&  \prod_{j,t} \text{Pr}(y_{j,t} | y_{j,<t}, s_{j,t}, a_j,c),\nonumber
\end{eqnarray}
\end{small}

\noindent where $c=\{u,i,r\}$ denotes the set of available context information.
Note that, in training, we have aspects and sketches available, and learn the model parameters by optimizing the joint probability in Eq.~\ref{eq-joint} over all the seen reviews. While, for test, the aspects and sketches are unknown. 
We  need to first infer an aspect sequence and then predict the corresponding sketch for each sentence. Finally, we generate the review content based on the predicted aspect and sketch information.

\section{The Proposed Approach}
\begin{figure}[!tb]
\centering 
\includegraphics[width=0.45\textwidth]{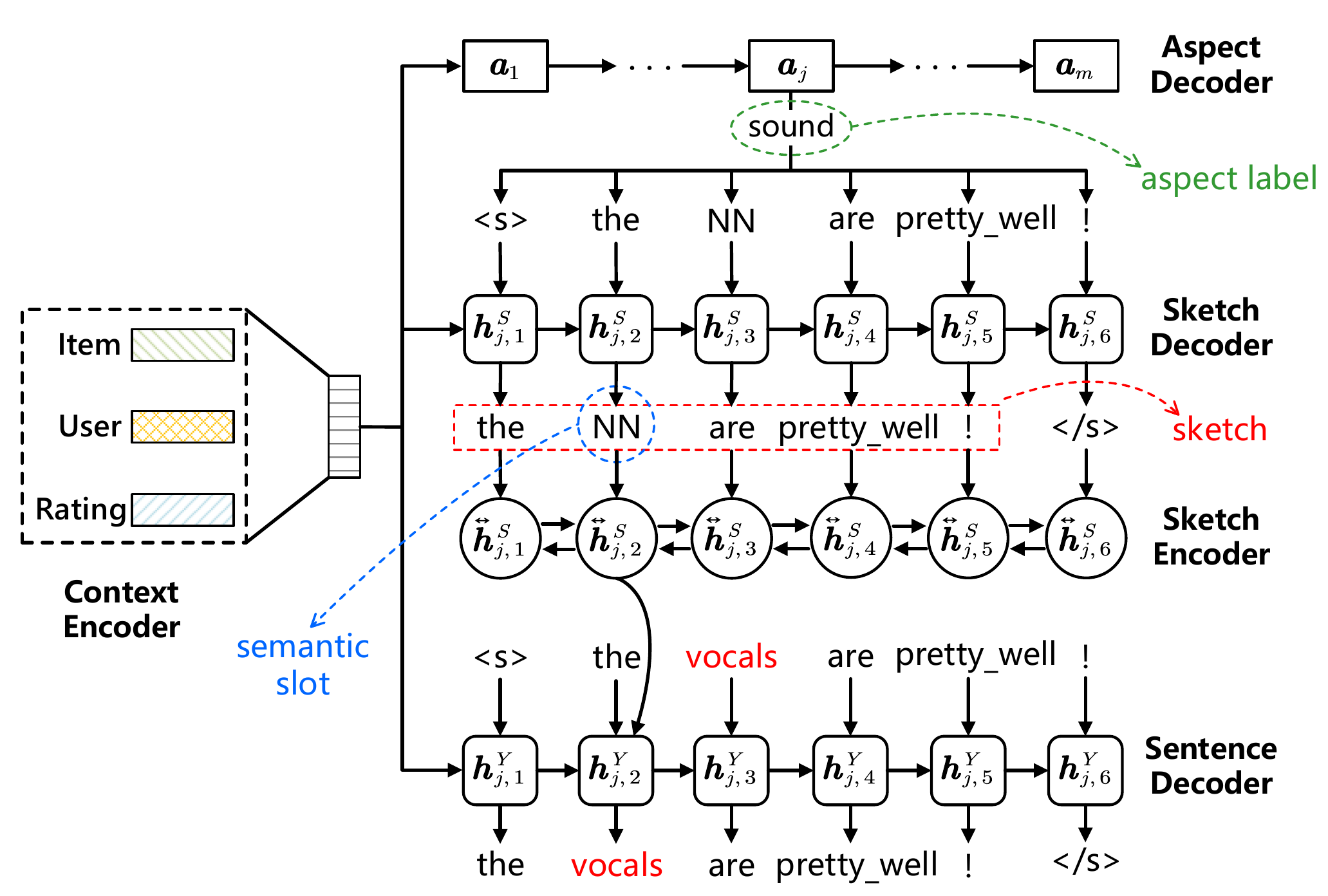} 
\caption{The overview of the proposed review generation model with the example of ``the vocals are pretty well". The predicted aspect label is \emph{sound}, and the generated sketch is ``the \underline{NN} are pretty\_well". } 
\label{fig-model} 
\end{figure}

Unlike previous works generating the review in a single stage,
we decompose the generation process into three stages, namely 
aspect sequence generation, aspect-aware sketch generation and sketch-based sentence generation. We present an overview illustration of the proposed model in Fig.~\ref{fig-model}.  Next we describe each part in detail.

\subsection{Aspect Sequence Generation}
To learn the model for generating aspect sequences, we need to 
derive the aspect sequence for training, and then decode the aspect sequence based on the context encoder. 

\paratitle{Aspect Extraction}. Aspects provide an informative summary about the feature or attribute information about a product or an item. For example, aspects of a restaurant may include food, staff and price, etc.
It is time-consuming and laborious to manually discover the aspects from texts. Here, we use an automatic unsupervised topic modeling approach to learning the aspects from the review content. 
Based on the Twitter-LDA model~\cite{ZhaoJWHLYL11}, we treat a review as a document consisting of multiple sentences. Each document is associated with a distribution over the aspects. When generating a sentence, an aspect label (or ID) is first sampled according to the document's distribution over the aspects. Then, the entire sentence is generated according to the word distribution conditioned on the aspect label. To purify the aspect words, we further incorporate 
a background language model to absorb background words. 
When topic models have been learned, we can derive a set of $A$ aspect-specific word distributions, denoted by $\{\theta_{\cdot}^{a}\}$, where $\theta_{w}^{a}$ denotes the probability of a word $w$ from the vocabulary $\mathcal{V}$ in aspect $a$.

\paratitle{Context Encoder}. Our aspect generation module adopts an encoder-decoder architecture. We first develop the context encoder based on the information of user $u$, item $i$ and rating score $r$. We first use a look-up layer to transform the three kinds of information into low-dimensional vectors.
Let $\bm{v}_u \in \mathbb{R}^{d_E}$, $\bm{v}_i \in \mathbb{R}^{d_E}$ and $\bm{v}_r \in \mathbb{R}^{d_E}$ denote the embeddings for $u$, $i$ and $r$ respectively. 
Then, we feed the concatenated vector into a Multi-Layer Perceptron (MLP) and produce a single vectorized representation $\bm{v}_c \in \mathbb{R}^{d_C}$:

\begin{equation}\label{eq-MLP}\small
\bm{v}_c= \text{MLP}([\bm{v}_u; \bm{v}_i; \bm{v}_r]).
\end{equation}
The embedding $\bm{v}_c$ summarizes the necessary information from the three kinds of context data. It is flexible to incorporate more kinds of useful information using a similar approach.

\paratitle{Aspect Decoder}. The decoder is built upon the GRU-based RNN network. Let $\bm{h}^A_j \in \mathbb{R}^{d_{H_A}}$ denote a $d_{H_A}$-dimensional hidden vector at the $j$-th time step, which is computed via:

\begin{equation}\small
\bm{h}^A_j = \text{GRU}(\bm{h}^A_{j-1}, \bm{v}_{a_{j-1}}),
\end{equation}
where $\bm{v}_{a_{j-1}} \in \mathbb{R}^{d_A}$ is the embedding of the previous aspect label $a_{j-1}$. The hidden vector of the first time step is initialized by the encoding vector $\bm{h}^A_0=\bm{v}_c$ in Eq.~\ref{eq-MLP}. Then, RNNs recurrently compute hidden vectors,  and predict the next aspect label (or ID) $a_{j}$. Additionally, we use an attention mechanism~\cite{LuongPM15} to enhance the effect of context information. We compute the attention score of context $c_k$ for the current time step  of the decoder via:

\begin{equation}\label{attn_score}\small
w^{(t)}_{k} = \frac{\exp(\tanh(\bm{W}_1 [\bm{h}^A_t;\bm{v}_{c_k}]))}{\sum_{c_{k'} \in \{u,i,r\}} \exp (\tanh(\bm{W}_1 [\bm{h}^A_t;\bm{v}_{c_{k'}}]))},   
\end{equation}
where $\bm{W}_1$ is the parameter matrix to learn, and the attention vector $\tilde{\bm{c}}_t$ is obtained by:

\begin{equation}\label{attn}\small
\tilde{\bm{c}}_t = \sum_{c_{k} \in \{u,i,r\}} w^{(t)}_{k}\bm{v}_{c_k}
\end{equation}

Finally, we compute the probability of the $j$-th aspect label $p(a_t|a_{<j},c)$ via:

\begin{small}
\begin{eqnarray}
\text{Pr}(a_j|a_{<j},c) &=& \text{softmax}(\bm{W}_4 \tilde{\bm{h}}^A_j+\bm{b}_1),  \label{eq-aspect_prob}   \\
\tilde{\bm{h}}^A_j  &=& \tanh(\bm{W}_2 \tilde{\bm{c}}_j +\bm{W}_3 \bm{h}^A_j),
\end{eqnarray}
\end{small}

\noindent where $\bm{W}_2$, $\bm{W}_3$, $\bm{W}_4$ and $\bm{b}_1$ are learnable parameter matrices or vector.

\subsection{Aspect-Aware Sketch Generation}
A sketch is a symbol sequence describing the skeleton of a sentence, where each symbol denotes a semantic symbol such as a POS tag or a bi-gram. 
Similar to the aspect decoder, we also use the GRU-based RNNs to implement the sketch decoder.
As shown in Fig.~\ref{fig-example}, the sketches \emph{w.r.t.} varying aspects are likely to be different. Hence, we need to consider the effect of aspect information in the generation of a sketch.  
 Let $\bm{h}_{j,t}^S \in \mathbb{R}^{d_{H_S}}$ denote a $d_{H_S}$-dimensional hidden vector at time step $t$ for the $j$-th sketch, which is computed via:

\begin{equation}\label{eq-hs}\small
\bm{h}_{j,t}^S = \text{GRU}(\bm{h}_{j,t-1}^S, \bm{x}^S_{j,t}),
\end{equation}
where $\bm{x}^S_{j,t}$ is further defined as 
\begin{equation}\small
\bm{x}_{j,t} = \bm{v}_{s_{j,t-1}}\odot \bm{v}_{a_j},
\end{equation} 
where $\bm{v}_{s_{j,t-1}} \in \mathbb{R}^{d_S}$ denotes the embedding for the previous  sketch symbol $s_{j,t-1}$,  $\bm{v}_{a_j}$ denotes the embedding of the current aspect, and ``$\odot$" denotes the element-wise product.
In this way, the aspect information can be utilized at each time step for generating an entire sketch.
We set the initial hidden vector for the $j$-th sketch as the last embedding of the previous sketch:
$\bm{h}_{j,0}^S=\bm{h}_{j-1,n'_{j-1}}^S$.
Specifically, we have $\bm{h}_{1,0}^S=\bm{v}_c$ for initialization.

Similar to Eq.~\ref{attn_score} and \ref{attn}, we can further use an attention mechanism for incorporating context information, and produce a context-enhanced sketch representation $\tilde{\bm{h}}_{j,t}^S$ for time step $t$. Finally, we compute $\text{Pr}(s_{j,t}|s_{j,<t},a_j,c)$ via:

\begin{equation}\label{eq-sketch_prob}\small
\text{Pr}(s_{j,t}|s_{j,<t},a_j,c)= \text{softmax} (\bm{W}_5 \tilde{\bm{h}}_{j,t}^S+\bm{W}_6 \bm{v}_{a_j}+ \bm{b}_2),
\end{equation}
where we incorporate the embedding $\bm{v}_{a_j}$ of the aspect $a_j$ for enhancing the aspect semantics.


\subsection{Sketch-based Review Generation}
When the aspect sequence and the sketches are learned, we can generate the word content of a review. Here, we focus on the generation process of a single sentence.

\paratitle{Sketch Encoder}. To encode the sketch information, we employ the a bi-directional GRU encoder~\cite{SchusterP97,ChoMGBBSB14} to encode the sketch sequence $s_{j, 1:n'_j}$ into a list of hidden vectors ${\{\overleftrightarrow{\bm{h}}^S_{j,t}\}}^{n'_j}_{t=1}$, where $\overleftrightarrow{\bm{h}}^S_{j,t}$ denotes the hidden vector for the $t$-th position in the $j$-th sketch at time step $t$ from the encoder. Different from Eq.~\ref{eq-hs}, we use a bi-directional encoder since the sketch is available at this stage, capturing the global information from the entire sketch.

\paratitle{Sentence Decoder}. Consider the word generation at time step $t$. Let $ \bm{v}_{y_{j,t-1}} \in \mathbb{R}^{d_Y}$ denotes the embedding of the previous word $y_{j,t-1}$.
As input, we concatenate the current sketch representation and  the embedding of the previous word
\begin{equation}\small
\bm{x}^Y_{j,t}=\overleftrightarrow{\bm{h}}^S_{j,t}\oplus \bm{v}_{y_{j,t-1}},
\end{equation}
where ``$\oplus$" denotes the vector concatenation.
Then, we  compute the hidden vector $\bm{h}^Y_{j,t} \in \mathbb{R}^{d_{H_Y}}$ for the $j$-th sentence via: 

\begin{equation}\small
\bm{h}^Y_{j,t} = \text{GRU}(\bm{h}^Y_{j,t-1},\bm{x}^Y_{j,t}).
\end{equation}
\ignore{where we concatenate the current sketch representation and  the embedding of the previous word
\begin{equation}\small
\bm{x}^Y_{j,t}=\overleftrightarrow{\bm{h}}^S_{j,t}\oplus \bm{v}_{y_{j,t-1}},
\end{equation}
where $ \bm{v}_{y_{j,t-1}} \in \mathbb{R}^{d_Y}$ denotes the embedding of word $y_{j,t-1}$.
}
Similar to Eq.~\ref{attn_score} and \ref{attn}, we further leverage the context to obtain an enhanced state representation denoted by $\tilde{\bm{h}}^Y_{j,t}$ using the attention mechanism. Then we transform it into an intermediate vector with the dimensionality of the vocabulary size:
\begin{equation}\small
\bm{z} = \tanh(\bm{W}_7[\tilde{\bm{h}}^Y_{j,t};\bm{v}_{s_{j,t}}]+\bm{b}_3),
\end{equation}
where  $\bm{v}_{s_{j,t}}$ is the embedding of the sketch symbol $s_{j,t}$.
By incorporating aspect-specific word distributions, we can apply the softmax function  to derive the generative probability of the $t$-th word

\begin{small}
\begin{equation}\label{eq-word_prob}
\text{Pr}(y_{j,t}|y_{j,<t},s_{j,1:n'_j},a_j,c) = \text{softmax}(z_{y_{j,t}}+ \theta^{a_j}_{y_{j,t}}),
\end{equation}
\end{small}

\noindent where $\theta^{a_j}_{y_{j,t}}$ is the probability from the word distribution for aspect $a_j$. Here, we boost the importance of the words which have large probabilities in the corresponding topic models.
In this process,  the generation of words is required to match the generation of sketch symbols slot by slot.
Here, we align  words and sketch symbols by using the same indices for each slot for ease of understanding. 
However,  the length of the sketch is not necessarily equal to that of the generated sentence, since 
a sketch symbol can correspond to a multi-term phrase. 
When the sketch token is a term or a phrase (\eg bi-grams), we directly copy the original terms or phases to the output slot(s).

\ignore{Finally, analogously to Equations \ref{sketch_attn} \textasciitilde \ref{sketch_prob}, the output probability for word $r_{i,t}$ is given by:
\begin{eqnarray}
\begin{split}
\textbf{x}^{attn}_t = \tanh (W_D \textbf{d}_{i,t} + W_R \textbf{c}^R_t   \\
                             + W_A \textbf{g}(a_i) + \textbf{b})   
\end{split}  \label{review_attn} \\ \notag  \\
p(r_{i,t}|r_{i,<t},\textbf{k}_i,a_i,c) = p_v(r_{i,t}) + \mathbbm{1}_{r_{i,t} \in a_i}  \label{review_prob}
\end{eqnarray}
where $\textbf{c}^R_t$ is the attention vector from context encoder, $p_v(r_{i,t})$ is the probability in vocabulary, and $\mathbbm{1}_{r_{i,t} \in A_i}$ is the probability  in aspect $a_i$. The initialized hidden vector $\textbf{d}_{i,0}$ is the summation of the $i$-th hidden vector of topic decoder and the last hidden vector of the $i-1$-th review decoder:
\begin{equation}
\textbf{d}_{i,0} = \textbf{h}_i + \textbf{d}_{i-1,l_y}
\end{equation}
The sketch constrains the decoding output. If the sketch word $k_{i,t}$ is glossed over, we use the hidden vector $\textbf{d}_{i,t}$ to compute $p(r_{i,t}|r_{i,<t}, \textbf{k}_i, \textbf{a}_i, c)$. In other cases, we force $r_{i,t}$ to conform to the sketch token $k_{i,t}$.
}

\subsection{Training and Inference}
Integrating Eq.~\ref{eq-aspect_prob}, \ref{eq-sketch_prob} and \ref{eq-word_prob} into Eq.~\ref{eq-joint}, we derive the joint model for review generation. 
We take the log likelihood of Eq.~\ref{eq-joint} over all training reviews as the objective function. 
The joint objective function is difficult to be directly optimized. 
Hence, we incrementally train the three parts, and fine-tune the shared or dependent
parameters in different modules with the joint objective. For training, we directly use the real aspects and sketches for learning the model parameters. For inference, we apply our model in a pipeline way: we  first infer the aspect,  then predict the sketches and finally generate the words using inferred aspects and sketches. 
During inference, for sequence generation, we apply the beam search method with beam size 4.


In the three sequence generation modules of our model, we incorporate two special symbols to indicate the start and end of a sequence, namely \textsc{Start} and \textsc{End}.
Once we generate the \textsc{End} symbol, the generation process will be stopped. 
Besides, we set the maximum generation lengths for aspect sequence and sketch sequence to be 5 and 50, respectively. 
In the training procedure, we adopt the Adam optimizer~\cite{KingmaB14}. 
In order to avoid overfitting, we adopt the dropout strategy with a rate of 0.2.
More implementation details can be found in Section 5.1 (see Table~\ref{tb-parameters}).

%% file: sec-exp.tex
\section{Experiments} \label{experiment}
In this section, we first set up the experiments, and then report the results and analysis.

\begin{table}
\centering
\small
\begin{tabular}{c| c |c |c |c}
\hline
Datasets &\#Users& \#Items & \#Reviews &\#Words \\
\hline\hline
\textsc{Amazon} & 89,672 & 31,829 & 681,004 & 22,570\\
\hline
\textsc{Yelp} & 95,617 & 37,112 & 1,063,420 & 31,861 \\
\hline
\textsc{RateBeer} & 12,266 & 51,365 & 2,487,369 & 42,757\\
\hline
\end{tabular}
\caption{Statistics of our datasets after preprocessing.} 
\label{tab:data}
\vspace{-0.3cm}
\end{table}

\subsection{Experimental Setup}

\paratitle{Datasets}. We evaluate our model on three real-world review datasets, including \textsc{Amazon} Electronic dataset~\cite{HeM16},  \textsc{Yelp} Restaurant dataset\footnote{https://www.yelp.com/dataset}, and \textsc{RateBeer} dataset~\cite{McAuleyLJ12}.
We convert all text into lowercase, and perform tokenization using NLTK\footnote{https://www.nltk.org}. We keep the words occurring at least ten times as vocabulary words.
We discard reviews with more than 100 tokens, and remove users and products (or items)  occurring fewer than five times. The reviews of each dataset are randomly split into training, validation and test sets (80\%/10\%/10\%). The detailed statistics of the three datasets are summarized in Table \ref{tab:data}.

\paratitle{Aspect and Sketch Extraction}. After the preprocessing, we use the Twitter-LDA model in \cite{ZhaoJWHLYL11} for automatically learning the aspects and aspect keywords. The numbers of aspects are set to 10, 5, and 5 for the three datasets, respectively. The aspect numbers are selected using the perplexity score on validation set. 
By inspecting into the top aspect words, we find the learned aspects are very coherent and meaningful.  
For convenience, we ask a human labeler to annotate each learned aspect from topic models with an \emph{aspect label}. 
Note that aspect labels are only for ease of presentation, and will not be used in our model. 
With topic models, we further tag each sentence with the aspect label which gives the 
maximum posterior probability conditioned on the words.
To derive the sketches, we first extract the most popular 200 bi-grams and tri-grams by frequency. We replace their occurrences with $n$-gram IDs. 
Furthermore, we keep the words ranked in top 50 positions of an aspect, and replace the occurrences of the rest words with their Part-of-Speech tags. 
We also keep the top 50 frequent words in the entire text collection, such as background words ``I" and ``am".  In this way, for each review, we obtain a sequence of aspect labels; for each sentence in the review, we  obtain a sequence of sketch symbols. Aspect sequences and sketch sequences are only available during the training process. 

\begin{table}
\centering
\begin{small}
\begin{tabular}{c | c}
\hline
Modules & Settings\\\hline\hline
Aspect & \tabincell{l}{$d_A=512$, $d_E=512$, $d_{H_A}=512$, \\ \#GRU-layer=$2$, 
batch-size=$1024$, \\ init.-learning-rate=$0.00002$, Adam optimizer}\\
\hline
Sketch & \tabincell{l}{$d_S=512$, $d_{H_S}=512$, \\ \#GRU-layer=$2$, 
batch-size=$64$,\\ init.-learning-rate=$0.0002$, \\ learning-rate-decay-factor=0.8, \\
learning-rate-decay-epoch=2, Adam optimizer}\\
\hline
Review & \tabincell{l}{$d_Y=512$, $d_{H_Y}=512$, \\ \#GRU-layer=$2$, 
batch-size=$64$,\\ init.-learning-rate=$0.0002$, \\ learning-rate-decay-factor=0.8, \\
learning-rate-decay-epoch=2, Adam optimizer}\\
\hline
\end{tabular}
\end{small}
\caption{Parameter settings of  the three modules in our model.} 
\label{tb-parameters}
\vspace{-0.3cm}
\end{table}

\paratitle{Baseline Models}. We compare our model against a number of baseline models:

\textbullet~\textmd{gC2S}~\cite{TangYCZM16}: It adopts an encoder-decoder architecture to generate review texts conditioned on context information through a gating mechanism.

\textbullet~\textmd{Attr2Seq}~\cite{ZhouLWDHX17}: It adopts an attention-enhanced attribute-to-sequence architecture to generate reviews with input attributes.

\textbullet~\textmd{TransNets}~\cite{Catherine2018}: It applies a student-teacher like architecture for review generation by representing the reviews of a user and an item into a text-related representation, which is regularized to be similar to the actual review's latent representation at training time.

\textbullet~\textmd{ExpansionNet}~\cite{NiM18}: It uses an encoder-decoder framework to generate personalized reviews by incorporating short phrases (\eg review summaries, product titles) provided as input and introducing aspect-level information (\eg aspect words).

\textbullet~\textmd{SeqGAN}~\cite{YuZWY17}: It regards the generative model as a stochastic parameterized policy and uses Monte Carlo search to approximate the state-action value. The discriminator is a binary classifier to evaluate the sequence and guide the learning of the generative model.

\textbullet~\textmd{LeakGAN}~\cite{GuoLCZYW18}: The generator is built upon a hierarchical reinforcement learning architecture, which consists of a high-level module and a low-level  module, and the discriminator is a CNN-based feature extractor. The advantage is that this model can generate high-quality long text by introducing the leaked mechanism.

 Among these baselines, gC2S, Attr2Seq and TransNets are context-aware generation models in different implementation approaches, ExpansionNet introduces external information such as aspect words, and SeqGAN and LeakGAN are GAN based text generation models. 
Original SeqGAN and LeakGAN are designed for general sequence generation without considering context information (\eg user, item, rating). 
The learned aspect keywords are provided as input for both ExpansionNet and our model.
All the methods have several parameters to tune. We employ validation set to optimize the parameters in each method. 
To reproduce the results of our model, we report the parameter setting used throughout the experiments in Table~\ref{tb-parameters}.
Our code is available at \url{https://github.com/turboLJY/Coarse-to-Fine-Review-Generation}.

\begin{table*}
\renewcommand\arraystretch{1.1}
\small
\begin{center}
\begin{tabular}{c||l||c c c c c c}
			\hline
			\textmd{Datasets} & \textmd{Models} & \textmd{Perplexity} & \textmd{BLEU-1(\%)} & \textmd{BLEU-4(\%)} & \textmd{ROUGE-1} & \textmd{ROUGE-2} & \textmd{ROUGE-L} \\
			\hline \hline
			\multirow{3}[6]{*}{\textsc{Amazon}}
			& gC2S & 38.67 & 24.14 & 0.85 & 0.262 & 0.046 & 0.212 \\
			& Attr2Seq & 34.67 & 24.28 & 0.88 & 0.263 & 0.043 & 0.214 \\
			& TransNets & 34.21& 21.61 & 0.60 & 0.227 & 0.026 & 0.199 \\
			& ExpansionNet & 31.50 & 26.56 & 0.95 & 0.290 & 0.052 & 0.262 \\
			\cline{2-8}
			& SeqGAN& 28.50 & 25.18 & 0.84 & 0.265 & 0.043 & 0.220 \\
			& LeakGAN & 27.66 & 25.66 & 0.92 & 0.267 & 0.050 & 0.236 \\
		   \cline{2-8}
			& Our model & \textbf{26.55} & \textbf{28.22} & \textbf{1.04} & \textbf{0.315} & \textbf{0.066} & \textbf{0.280} \\
			\hline \hline
			\multirow{3}[6]{*}{\textsc{Yelp}}
			& gC2S & 35.52 & 24.39 & 0.87 & 0.243 & 0.046 & 0.188 \\
			& Attr2Seq & 33.12 & 24.71 & 0.89 & 0.245 & 0.047 & 0.191 \\
			& TransNets & 34.81 & 21.41 & 0.35 & 0.202 & 0.026 & 0.156 \\
			& ExpansionNet & 29.53 & 27.46 & 1.06 & 0.276 & 0.061 & 0.216 \\
			\cline{2-8}
			& SeqGAN & 26.84 & 24.83 & 0.99 & 0.253 & 0.054 & 0.192 \\
			& LeakGAN & 25.53 & 25.96 &  1.03 & 0.271 & 0.056 & 0.208 \\
			\cline{2-8}
			& Our model & \textbf{23.96} & \textbf{29.43} & \textbf{1.13} & \textbf{0.284} & \textbf{0.070} & \textbf{0.235} \\
			\hline \hline
			\multirow{3}[6]{*}{\textsc{RateBeer}}
			& gC2S & 17.81 & 32.13 & 5.55 & 0.379 & 0.140 & 0.331 \\
			& Attr2Seq & 16.84 & 32.21 & 5.80 & 0.380 & 0.142 & 0.331 \\
			& TransNets & 19.08 & 29.74 & 3.61 & 0.347 & 0.114 &0.302 \\
			& ExpansionNet & 17.07 & 34.53 & 6.83 & 0.400 & 0.156 & 0.376 \\
			\cline{2-8}
			& SeqGAN & 14.30 & 32.41 & 5.62 & 0.369 & 0.146 & 0.337 \\
			& LeakGAN & 13.74 & 33.76 & 6.03 & 0.378 & 0.142 & 0.355 \\
			\cline{2-8}
			& Our model & \textbf{13.07} & \textbf{36.11} & \textbf{7.04} & \textbf{0.422} & \textbf{0.164} & \textbf{0.393} \\
			\hline
\end{tabular}
\caption{Performance comparisons of different methods for automatic review generation using three datasets.}
\label{tab:main-results}
\end{center}
\vspace{-0.3cm}
\end{table*}

\paratitle{Evaluation Metrics}. To evaluate the performance of different methods on automatic review generation, 
we  adopt six evaluation metrics, including Perplexity, BLEU-1/BLEU-4, ROUGE-1/ROUGE-2/ROUGE-L. Perplexity\footnote{https://en.wikipedia.org/wiki/Perplexity} is the standard measure for evaluating language models; BLEU~\cite{PapineniRWZ02} measures the ratios of the co-occurrences of $n$-grams between the generated and real reviews;
ROUGE~\cite{Lin04} measures the review quality by counting the overlapping $n$-grams between the generated and real reviews.


\subsection{Results and Analysis}
In this subsection, we construct a series of experiments on the effectiveness of the proposed model for the review generation task. 

\paratitle{Main Results}. Table~\ref{tab:main-results} presents the performance of different methods on automatic review generation. We can make the following observations. First, among the three context-based baselines, gC2S and Attr2Seq perform better than TransNets. 
The two models have similar network architectures, which are simpler than TransNets. 
We find they are easier to obtain a stable performance on large datasets. 
Second, GAN-based methods work better than the above baselines, especially LeakGAN. 
LeakGAN is specially designed for generating long text, and we adapt it to our task by incorporating context information. Third, ExpansionNet performs best among all the baseline models. A major reason is that it incorporates external knowledge such as review summaries, product titles and aspect keywords. 
Finally, our model outperforms all the baselines with a large margin. 
These baseline methods perform the generation in a single stage. 
As a comparison, we use a multi-stage  process to gradually generate long and informative reviews in a coarse-to-fine way. Our model is able to better utilize aspect semantics and syntactic sketch, which is the key of the performance improvement over baselines. 
Overall, the three datasets show the similar findings. In what follows, we will report the results on \textsc{Amazon} data due to space limit. 
We select the best two baselines \emph{ExpansionNet}  and \emph{LeakGAN} as reference methods.  

\begin{table}[t]
	\small
	\centering
	\begin{tabular}{ l || c c  }
		\hline
		Models & BLEU-1(\%) & ROUGE-1 \\
		\hline
		\hline
		Our model & 28.22 & 0.315 \\
		\hline
		w/o aspect & 27.85 & 0.296 \\
		w/o sketch & 25.95 & 0.273 \\
		\hline
	\end{tabular}
	\caption{Ablation analysis on \textsc{Amazon} dataset.} 
	\label{tab:ablation-results}
	\vspace{-0.3cm}
\end{table}

\paratitle{Ablation Analysis}. The major novelty of our model is that it incorporates 
two specific modules to generate aspects and sketches respectively. 
To examine the contribution of the two modules, we compare our model with its two variants by removing either of the two modules. 
We present the BLEU-1 and ROUGE-1 results of our model and its two variants  in Table \ref{tab:ablation-results}. As we can see, both components are useful to improve the final performance, and the sketch generation module seems more important in our task. 
In our model, the aspect generation module is used to cover aspect semantics and generate informative review;  the sketch generation module is able to utilize syntactic templates to improve the generation fluency, especially for long sentences. 
Current experiments evaluate the usefulness of the two modules based on the overall generation quality. Next, we verify their functions using two specific experiments, namely aspect coverage and fluency evaluation.


\paratitle{Aspect Coverage Evaluation}. A generated review  is informative if it can effectively capture the semantic information of the real review. 
Following~\cite{NiM18}, we examine the aspect coverage of different models.  
Recall that we have used topic models to tag each sentence with an aspect label (or ID). 
We analyze the average number of aspects in real and generated reviews, and compute on average how many aspects in real reviews are covered in generated reviews. We consider a review as covering an aspect if any of the top 50 words of an aspect exists in the review\footnote{For accuracy, we manually remove the irrelevant words (about 5\%$\sim$10\%) from the top 50 words in each aspect. }.
In Table \ref{tab:aspect-results}, we first see an interesting observation that LeakGAN is able to generate more aspects but yield fewer real aspects.
As a comparison, ExpansionNet and our model perform better than LeakGAN by covering more real aspects, since the two models use the aspect information to instruct the review generation. Our model is better than ExpansionNet by characterizing the aspect transition sequences.
These results indicate the usefulness of the aspect generation module in capturing more  semantic information related to a review.

\paratitle{Fluency Evaluation}. We continue to evaluate the usefulness of the sketch generation module  in improving the fluency of the generated text.  Following~\cite{XuRL018}, we construct the fluency evaluation to examine how likely the generated text is produced by human.  We randomly choose 200 samples from test set. A sample contains the input contexts (\ie user, item, rating), and the texts generated by different models. 
It is difficult to develop automatic evaluation methods for accurate fluency evaluation. Here, we invite two human annotators (excluding the authors of this paper) who have good knowledge in the domain of electronic reviews to assign scores to the generated reviews.
They are required to assign a score to a generated (or real) review according to a 5-point Likert scale\footnote{https://en.wikipedia.org/wiki/Likert\_scale} on fluency. 
In the 5-point Likert scale, 5-point means ``very satisfying'', while 1-point means ``very terrible''.
We further average the two annotated scores over the 200 inputs. 
The results are shown in Table \ref{tab:human-results}. We can see that our model 
achieves the highest fluency score among the automatic methods. 
By using sketches, our model is able to leverage the learned syntactic patterns from available  reviews. 
The Cohen's kappa coefficients are above 0.7, indicating a high correlation and agreement between the two human annotators.


\begin{table}[t]
	\centering\small
	\begin{tabular}{ l || l  l  l }
		\hline
		Models & \tabincell{l}{\# aspects\\(real)} & \tabincell{l}{\# aspects\\(generated)} & \tabincell{l}{\# covered\\aspects} \\
		\hline
		\hline
		ExpansionNet & 2.41 &2.02 & 0.885 \\
		LeakGAN & 2.41 & \textbf{2.18} & 0.630 \\
		Our model & 2.41 & 2.03 & \textbf{1.076} \\
		\hline
	\end{tabular}
	\caption{Aspect coverage evaluation on \textsc{Amazon} dataset.} 
	\label{tab:aspect-results}
	\vspace{-0.3cm}
\end{table}

\begin{table}[t]
	\centering
	\small
	\begin{tabular}{c || c  c  c c }
		\hline
		Measures & Gold &  ExpansionNet & LeakGAN & Our\\
		\hline
		\hline
		Fluency & 4.01 & 3.29 & 3.26 & \textbf{3.54} \\
		Kappa & 0.80 & 0.72 & 0.76 & 0.74 \\
		\hline
	\end{tabular}
	\caption{Fluency evaluation on  \textsc{Amazon} dataset. } 
	\label{tab:human-results}
	\vspace{-0.3cm}
\end{table}

\begin{table*}
	\begin{center}
		\begin{scriptsize}
			\begin{tabular}{c || c || c}
				\hline
				\textmd{Gold Standard} & \textmd{Generated Sketch} & \textmd{Generated Review} \\
				\hline \hline	
				\tabincell{l}{ ${\text{the shipping was quick and easy}}_{\textcolor[RGB]{255,0,0}{\text{service}}}$very good\\ $\text{product at a \underline{reasonable price} }_{\textcolor[RGB]{0,0,255}{\text{price}}}$ 5mm male to \\ $\text{2 rca stereo audio \underline{cable} }_{\textcolor[RGB]{	34,139,34}{\text{sound}}}$ highly \underline{recommend}  \\ $\text{this product to anyone}_{\textcolor[RGB]{139,69,19}{\text{overall}}}$ }  &
				
				\tabincell{l}{ ${\text{this cable worked\_perfectly for my NNS}_{\textcolor[RGB]{34,139,34}{\text{sound}}}}$\\ the price was very JJ and i would\_purchase \\ $\text{NN from this NN}_{\textcolor[RGB]{0,0,255}{\text{price}}}$ it VBD on\_time and\\ $\text{in good NN}_{\textcolor[RGB]{255,0,0}{\text{service}}}$ $\text{i would\_recommend it}_{\textcolor[RGB]{139,69,19}{\text{overall}}}$ } &
				
				\tabincell{l}{ ${\text{this \underline{cable} worked perfectly for my needs}_{\textcolor[RGB]{34,139,34}{\text{sound}}}}$ the\\ \underline{price} was very \underline{reasonable} and i would purchase\\ another $\text{from this vendor}_{\textcolor[RGB]{0,0,255}{\text{price}}}$ it arrived on time and\\ in good $\text{condition}_{\textcolor[RGB]{255,0,0}{\text{service}}}$ $\text{i would \underline{recommend} it}_{\textcolor[RGB]{139,69,19}{\text{overall}}}$ } \\
				\hline
				\hline	
				\tabincell{l}{ oxtail was good other than the flavors were very\\\underline{bland} $_{\textcolor[RGB]{255,0,0}{\text{food}}}$ \underline{place} is small so if the tables are full\\ $\text{be prepared to wait}_{\textcolor[RGB]{0,0,255}{\text{place}}}$ pay too much for what\\ $\text{you get}_{\textcolor[RGB]{34,139,34}{\text{price}}}$ $\text{i will not be back to this location}_{\textcolor[RGB]{139,69,19}{\text{overall}}}$ } &
				
				\tabincell{l}{ ${\text{i had the NN NN and it was very JJ}_{\textcolor[RGB]{255,0,0}{\text{food}}}}$ the\\ $\text{staff was JJ but service was a little JJ}_{\textcolor[RGB]{238,48,167}{\text{service}}}$ i\\ $\text{had a bad\_experience at this NN}_{\textcolor[RGB]{0,0,255}{\text{place}}}$ i VBP\\ $\text{not JJ if i will be back RB}_{\textcolor[RGB]{139,69,19}{\text{overall}}}$ } &
				
				\tabincell{l}{${\text{i had the falafel wrap and it was very \underline{bland} }_{\textcolor[RGB]{255,0,0}{\text{food}}}}$ the\\ staff $\text{was friendly but service was a little slow}_{\textcolor[RGB]{238,48,167}{\text{service}}}$\\ $\text{i had a bad\_experience at this \underline{place} }_{\textcolor[RGB]{0,0,255}{\text{place}}}$ i am not\\ sure if i will $\text{be back again}_{\textcolor[RGB]{139,69,19}{\text{overall}}}$ } \\
				\hline
				\hline
				\tabincell{l}{ $\text{the \underline{aroma} is insanely sour from bad hops}_{\textcolor[RGB]{255,0,0}{\text{aroma}}}$\\ dark clear ruby red beat sugar \underline{flavor} and strong\\ $\text{\underline{alcohol} in aftertaste}_{\textcolor[RGB]{0,0,255}{\text{flavor}}}$ golden \underline{body} with a small\\ $\text{\underline{white head} }_{\textcolor[RGB]{34,139,34}{\text{body}}}$ ${\text{dont waste your money on this}}_{\textcolor[RGB]{139,69,19}{\text{overall}}}$} & 
				
				\tabincell{l}{${\text{VBZ an amber\_body with a JJ NN head}_{\textcolor[RGB]{34,139,34}{\text{body}}}}$\\ $\text{the flavor is very JJ with notes of NN}_{\textcolor[RGB]{0,0,255}{\text{flavor}}}$\\ this beer has the JJS aroma of canned\_corn\\ $\text{i have ever VBN}_{\textcolor[RGB]{255,0,0}{\text{aroma}}}$ } &
				
				\tabincell{l}{${\text{pours an amber \underline{body} with a \underline{white} finger \underline{head}}_{\textcolor[RGB]{34,139,34}{\text{body}}}}$\\ $\text{the \underline{flavor} is very horrible with notes of \underline{alcohol} }_{\textcolor[RGB]{0,0,255}{\text{flavor}}}$\\ this beer has the worst \underline{aroma} of canned corn i\\ $\text{have ever smelled}_{\textcolor[RGB]{255,0,0}{\text{aroma}}}$ } \\
				\hline
			\end{tabular}
			\caption{Samples of the generated reviews by our model. 
				The three reviews with rating scores of 5 (positive), 3 (neutral), and 1 (negative) are from \textsc{Amazon}, \textsc{Yelp} and \textsc{RateBeer} datasets, respectively. 
				For privacy, we omit the UIDs and PIDs. 
				For ease of reading, colored aspect labels are manually created corresponding to the predicted aspect IDs by our model. We have underlined important overlapping terms between real and generated reviews.  }
			\label{tab:examples}
		\end{scriptsize}
	\end{center}
	\vspace{-0.3cm}
\end{table*}

\subsection{Qualitative Analysis}
In this part,  we perform the  qualitative analysis on the quality of the generated reviews. 
We present three sample  reviews generated by our model in Table~\ref{tab:examples}.
As we can see, our model has  covered most of the major aspects (with many overlapping aspect keywords) of the real reviews. 
Although some generated sentences do not follow the exact syntactic structures of real reviews, they are very readable to users.
Our model is able to generate aspect-aware  sketches, which are very helpful to instruct the generation of the word content. 
With the aspect and sketch generation modules, our model is able to produce informative reviews consisting of multiple well-structured sentences. Another interesting observation is that the polarities of the generated text also correspond to their real rating scores, since the rating score has been modeled in the context encoder.


%% file: sec-con.tex
\section{Conclusion}
This paper presented a novel review generation model using an aspect-aware coarse-to-fine generation process. Unlike previous methods, our model decomposed the generation process into three stages focusing on different goals.
We constructed extensive experiments on three real-world review datasets. The results have demonstrated the effectiveness of our model in terms of overall generation quality, aspect coverage, and fluency. 
As future work, we will consider integrating more kinds of syntactic features from linguistic analysis such as dependency parsing.

%% file: sec-ack.tex
\section*{Acknowledgments}

This work was partially supported by the National Natural Science Foundation of China under Grant No. 61872369 and 61832017, the Fundamental Research Funds for the Central Universities, the Research Funds of Renmin University of China under Grant No. 18XNLG22 and 19XNQ047.